\documentclass[final]{IEEEtran}

\usepackage[numbers,sort&compress]{natbib}
\usepackage[colorlinks,urlcolor=blue,linkcolor=blue,citecolor=blue]{hyperref}

\usepackage{color,array}
\usepackage{booktabs}
\usepackage{graphicx}
\usepackage{amsmath}
\usepackage{multirow}
\usepackage{makecell}

\usepackage{pifont}
\usepackage[ruled,linesnumbered]{algorithm2e}

\begin{document}

\title{
Improving Quantization-aware Training of Low-Precision Network via Block Replacement on Full-Precision Counterpart} 

\author{%
  Chengting Yu$^{1,2}$, Shu Yang$^{2}$, Fengzhao Zhang$^{2}$, Hanzhi Ma$^{1,2}$, Aili Wang$^{1,2,*}$ Er-Ping Li$^{1,2}$  \\
  $^1$ College of Information Science and Electronic Engineering, Zhejiang University\\
  $^2$ ZJU-UIUC Institute, Zhejiang University\\
  chengting.21@intl.zju.edu.cn, ailiwang@intl.zju.edu.cn
}

\maketitle

\begin{abstract}
Quantization-aware training (QAT) is a common paradigm for network quantization, in which the training phase incorporates the simulation of the low-precision computation to optimize the quantization parameters in alignment with the task goals.
However, direct training of low-precision networks generally faces two obstacles: 1. The low-precision model exhibits limited representation capabilities and cannot directly replicate full-precision calculations, which constitutes a deficiency compared to full-precision alternatives; 2. Non-ideal deviations during gradient propagation are a common consequence of employing pseudo-gradients as approximations in derived quantized functions.
In this paper, we propose a general QAT framework for alleviating the aforementioned concerns by permitting the forward and backward processes of the low-precision network to be guided by the full-precision partner during training.
In conjunction with the direct training of the quantization model, intermediate mixed-precision models are generated through the block-by-block replacement on the full-precision model and working simultaneously with the low-precision backbone, which enables the integration of quantized low-precision blocks into full-precision networks throughout the training phase. 
Consequently, each quantized block is capable of: 1. simulating full-precision representation during forward passes; 2. obtaining gradients with improved estimation during backward passes.
We demonstrate that the proposed method achieves state-of-the-art results for 4-, 3-, and 2-bit quantization on ImageNet and CIFAR-10.
The proposed framework provides a compatible extension for most QAT methods and only requires a concise wrapper for existing codes.

\end{abstract}



\section{Introduction}
\label{sec:intro}

\IEEEPARstart{C}onvolutional neural networks (CNNs) have made substantial advancements in image classification \cite{he2016deep,krizhevsky2012imagenet}, semantic segmentation \cite{girshick2014rich,he2017mask}, object detection \cite{redmon2016you,ren2015faster}, and image restoration \cite{guo2017network}. With the large computational complexity of CNNs, there is an increasing demand for strategies to successfully deploy networks on resource-constrained devices with limited processing capability. Several strategies have been developed to train fast and compact neural networks that are both efficient in terms of speed and size, such as manually-designed architectures \cite{cui2019fast,howard2019searching,sandler2018mobilenetv2,zhang2018shufflenet}, pruning \cite{li2016pruning,frankle2018lottery,liu2018rethinking}, quantization \cite{jacob2018quantization,han2021improving,banner2019post,esser2020learned,li2019additive}, knowledge distillation \cite{hinton2015distilling,romero2014fitnets,wang2021knowledge}, and neural architecture search.  In this paper, we focus on network quantization, which attempts to transform the weights and/or activations of a full-precision (FP) network into low-precision (LP) approximations so that fixed-point computations can be conducted without compromising too much accuracy.

Quantization-aware training (QAT) has become one of the most effective methods for achieving low-bit quantization while maintaining relatively high accuracy by simulating low-precision operators during training and optimizing quantizer parameters with task objectives \cite{krishnamoorthi2018quantizing,esser2020learned,nahshan2021loss}.
The main difficulty in directly training a quantized network arises from the discretization stage, in which a discrete quantizer (i.e., scaled round function) converts a normalized value to one of the discrete levels.
First, since a discrete quantizer is non-differentiable, the straight-through estimator (STE) \cite{bengio2013estimating} is often utilized during backpropagation \cite{lee2021network,choi2018pact,esser2020learned,jung2019learning,park2020profit,zhang2018lq,zhou2016dorefa}. Despite recent STE-based methods demonstrating reasonable performance \cite{lee2021network,bhalgat2020lsq+,esser2020learned,jung2019learning,park2020profit}, STE may produce an unexpected gradient mismatch issue \cite{lee2021network,yang2019quantization} with distinct forward and backward passes, resulting in an undesirable drop in accuracy \cite{zhuang2020training}.
Second, due to the limited representation capability of the discrete quantizer \cite{kim2019qkd}, the accuracy of extremely low-bit quantized networks (i.e., 4-, 3-, or 2-bit) is inevitably diminished. In order to enhance the low-bit representation, certain methods are suggested, known as non-uniform quantization \cite{miyashita2016convolutional,zhou2017incremental,li2019additive,yamamoto2021learnable,liu2022nonuniform}. This technique entails modifying the quantization resolution in accordance with the density of the real-valued distribution in order to find suitable correspondences with the activation and weight distributions \cite{yamamoto2021learnable,li2019additive}. However, the increased inference burden presents them with major obstacles.

By consulting full-precision models for guidance as auxiliary supervision, optimal schemes are proposed to partially compensate for the inherent issues of QAT \cite{kim2019qkd,zhuang2020training,zhu2023quantized} through the incorporation of additional losses into intermediate layers to address the gradient approximation problem and provide regularization.
In this regard, the simplest approach commonly employed in QAT is to initialize the low-precision model directly with the weights obtained from the pre-trained full-precision model \cite{esser2020learned,li2019additive,bhalgat2020lsq+}. 
Besides, training strategies such as knowledge distillation have been combined to learn a low-precision student network by distilling knowledge from a full-precision teacher \cite{polino2018model,kim2019qkd, zhu2023quantized}; additionally, some strategies use full-precision auxiliary routines for dealing with quantizer noise \cite{zhuang2020training,boo2021stochastic}.

\begin{figure*}[t]
\centering
\includegraphics[width=\textwidth]{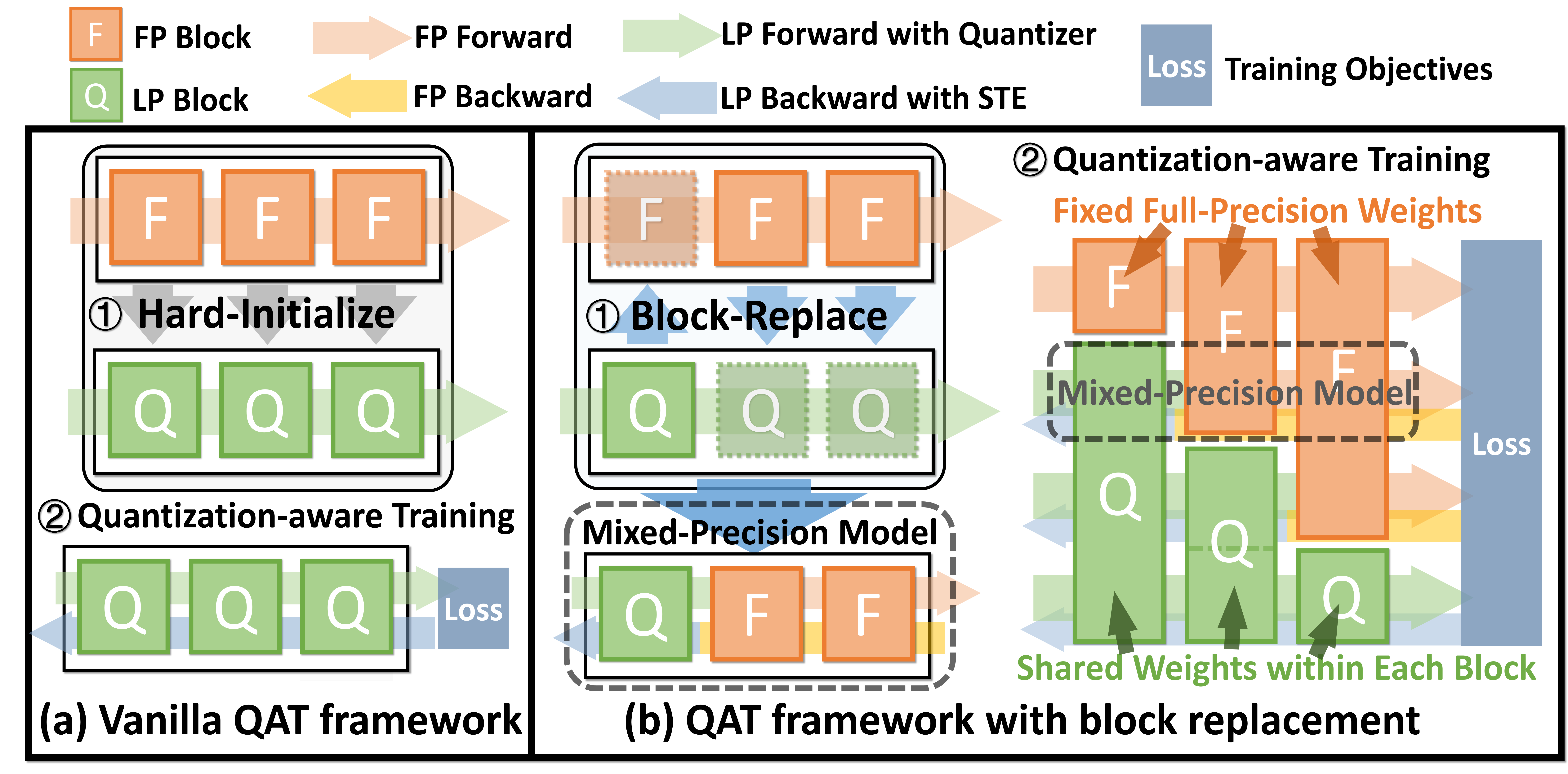} %
\caption{Framework Overview. (a) The fundamental implementation of Quantization-aware Training (QAT) in which weight initialization is performed using full-precision counterparts. (b) The proposed block-wise replacement framework (BWRF) generates mixed-precision models during the training phase, employing full-precision counterparts for auxiliary supervision.}
\label{fig1}
\end{figure*}

In this paper, we propose a novel, neat, and efficient framework for QAT methods to utilize the guiding potential of full-precision counterparts properly.
As shown in Fig. \ref{fig1}, instead of the straight and rough initialization, we build the mixed-precision models by converting the end FP blocks into LP blocks and then integrating those intermediate models into the entire training framework for implicit guidance.
The weights of FP blocks are fixed (not changed during training), while the weights of LP blocks learn from both FP- and LP-backward routines. 
By grafting the LP model into its FP counterpart, the proposed framework allows the front LP blocks to pass the end FP blocks to ensure the reliability of backward gradients, as well as making the LP representation mirror the FP representation to accommodate the following FP feature extractor. 
The aforementioned concerns of both forward and backward passes are addressed in this case. Empirical experiments support our hypothesis: 1. Each intermediate model outperforms the vanilla LP model, and the performance tendency decreases from the top (pure FP model) to the bottom (pure LP model); 2. The guidance from intermediate models is positive and valid, leading FP blocks to converge toward training targets; 
3. During training, the exchanging LP blocks gradually mimic their FP counterparts, executing an implicit regularization for FP representation. 
We validate the proposed framework based on uniform quantization \cite{esser2020learned}, and demonstrate its improved performance over the state-of-the-art methods for low-bit network quantization. 
\\
\noindent
Our contributions can be summarized as follows:
\begin{itemize}
    \item We propose a neat framework for QAT that makes complete use of a pre-trained, full-precision model. Our approach is capable of producing high-performance, low-bit quantization models, without increasing the model complexity at the inference phase.
    \item We evaluate the proposed framework through empirical observations, and uncover the framework's implicit insights in conjunction with our concerns and hypotheses.
    \item We demonstrate the effectiveness of our method under low-precision of 4-, 3-, and 4-bit widths, achieving state-of-the-art results on ImageNet.
\end{itemize}

\section{Related Work}

\noindent
\textbf{Network Quantization.} \cite{han2021improving} presents a detailed description of CNN quantization and divides quantizer designs into two categories: post-training quantization (PTQ) and quantization-aware training (QAT). PTQ techniques often quantize a network without complete training of full data \cite{banner2019post,choukroun2019low,finkelstein2019fighting,hubara2020improving,nagel2019data,zhao2019improving}. QAT techniques commonly outperform PTQ for low-precision networks with proper training on original data \cite{bhalgat2020lsq+,choi2018pact,esser2020learned,gong2019differentiable,jain2020trained,jung2019learning,zhou2016dorefa}. However, non-idealities appear in both forward and backward owing to the discrete quantizers used by the QAT during training. Previous QAT approaches have been investigated to optimize the quantization parameters (e.g., clipping value, quantization step size) \cite{esser2020learned,bhalgat2020lsq+} and introduce non-uniform quantization \cite{miyashita2016convolutional,zhou2017incremental,li2019additive,liu2022nonuniform,yamamoto2021learnable} in order to improve the representation of low-precision and effectively balance quantization error. Another line of QAT studies has attempted to tackle the gradient mismatch issue induced by the non-differentiable quantizer during QAT \cite{lee2021network,zhuang2020training}.
\\
\noindent
\textbf{Auxiliary supervision.} Auxiliary supervision seeks to improve network training by integrating auxiliary objectives and extra losses into intermediate layers, hence combating the gradient issue and providing regularization \cite{szegedy2015going,louizos2018relaxed,nekrasov2019fast,zhao2017pyramid}. \cite{zhuang2020training} was the first to suggest using full-precision branches to address the gradient problem in QAT, demonstrating the practicality of auxiliary supervision in QAT.
Knowledge distillation is a common approach for network training \cite{hinton2015distilling}, in which smaller networks are trained by transferring features from stronger teacher models, and it may also be thought of as auxiliary supervision. 
\cite{polino2018model} were the first to propose incorporating distillation into QAT, with the full-precision model serving as the teacher model and distillation losses established on the low-precision model's output layer for auxiliary supervision. Following research, such as QKD \cite{kim2019qkd} and QFD \cite{zhu2023quantized}, has enhanced the design of distillation branches for auxiliary supervision in QAT.
We develop full-precision branches for auxiliary supervision in the proposed quantization framework by grafting the full-precision model to better address the quantizer's gradient mismatch. 
Notably, our framework employs KD loss as one of the auxiliary learning sources based only on pre-trained FP counterparts to aid optimization, removing the need to train an extra teacher network. On this premise, we continue to outperform KD approaches without the assistance of large teachers.

\section{Approach}

As shown in Fig. \ref{fig1} (b), we propose a general technique for enhancing quantized-aware training (QAT) in this section: the block-wise replacement framework (BWRF). Initially, in accordance with depth and resolution, the identical network structure of the low-precision (LP) model and its full-precision (FP) counterpart is uniformly partitioned into multiple block-wise sections. For instance, the block sections of the ResNets family correspond to the code implementation of res-blocks. Grafted mixed-precision (MP) models are obtained by progressively replacing the last low-precision blocks with pre-trained full-precision blocks. 
Throughout the training phase, only the low-precision component of the mixed-precision model is modified; the full-precision component remains constant.
The parameters contained within individual LP blocks are synchronized and shared. 
Mixed-precision models provide full-precision branches that serve as auxiliary supervision for the low-precision backbone, which are neglected during the inference process. 
The primary objective of developing full-precision branches is to enhance the utilization of full-precision counterparts for guidance and to mitigate issues associated with limited representation in the forward pass and gradient estimation in the backward pass.

\subsection{Quantization-aware Training}

The fundamental concept underlying Quantization-aware Training (QAT) is to incorporate quantization simulation during network training, thus enabling the direct determination of the model's optimal solution while subject to quantization constraints. Floating-point activation and weights are converted to a restricted low-bits representation using quantizers during the forward pass. The quantizer, denotes as $q(\cdot)$, accepts an input vector $v$ and generates a pseudo-quantized output $\hat{v}$, as follows:
\begin{equation}
    \hat{v}=q(v)=s \cdot \lfloor clip(\frac{v}{s}, N, P) \rceil
    \label{eq:quantizer}
\end{equation}
where notation $\lfloor \cdot \rceil$ denotes the round-to-nearest operator, $s$ is the scaling factor, $clip(\cdot, N, P)$ represents the clipping function whose lower and upper quantization thresholds $N$ and $P$, respectively. 

During the backward pass, QAT encounters the non-differentiability of the round function in Eq. \ref{eq:quantizer}, which restricts gradient-based training.  A commonly used approach to address the issue is to approximate the true gradient for quantizers using the straight-through estimator (STE) \cite{bengio2013estimating,hinton2012neural}, which simply approximates the gradient of round function as:
\begin{equation}
    \frac{\partial}{\partial x} \Big( \lfloor x \rceil \Big) = 1
\end{equation}

\noindent
Then, the gradient passing quantizers can be obtained: 
\begin{equation}
\frac{\partial \hat{v}}{\partial v} = \begin{cases}
1, & \text{if } N < v < P; \\
0, & \text{otherwise}. 
\end{cases}
\label{eq:ste}
\end{equation}

\noindent
We follow LSQ \cite{esser2020learned} and APoT \cite{li2019additive} in practice, which allow scale parameters of quantizers to be learned; the gradient of $s$ can be derived from Eq. \ref{eq:quantizer} and Eq. \ref{eq:ste}: 
\begin{equation}
\frac{\partial \hat{v}}{\partial s} =  \begin{cases}
-v/s+\lfloor v/s \rceil, & \text{if } N < v/s < P; \\
N, & \text{if } v/s \leq N; \\
P,  & \text{if } v/s \geq P.
\end{cases} 
\end{equation}

\noindent
In accordance with the standard configuration of QAT, the quantizers for activations and weights are established prior to the linear operators (e.g., convolutional and fully-connected layers) to ensure that matrix multiplication can occur in fix-point domains \cite{choi2018pact,li2019additive,esser2020learned}.
During code implementation, the functionalities of quantizers are encapsulated within linear operators to facilitate their efficient integration into the structure of the corresponding FP model.

\begin{figure}[t]
\centering
\includegraphics[width=1.0\columnwidth]{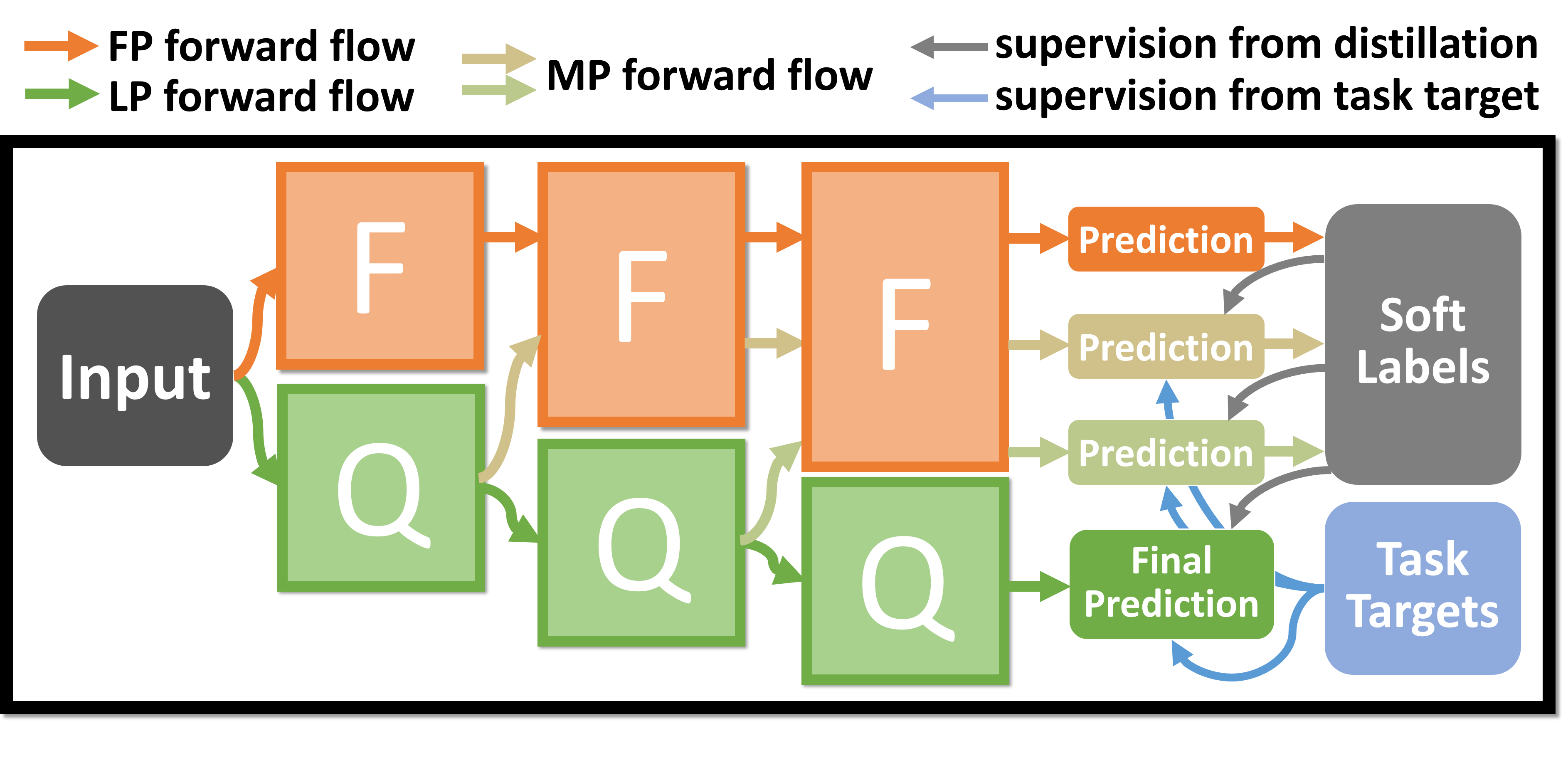} 
\caption{Implementation of BWRF Training.
Mixed-precision models are implemented implicitly through the utilization of overlapping LP forward flows. 
Both task targets and model predictions are regarded as loss sources for training.}
\label{fig2}
\end{figure}

\subsection{Framework Formulation}
Consider the entire definition of the framework. Given the LP model $Q$ and its corresponding FP model $F$,
\begin{equation}
    Q=\{Q_1,Q_2,…,Q_n\}, F=\{F_1,F_2,…,F_n\}
\end{equation}
where $n$ represents the quantity of blocks, $Q_i$ and $F_i$ denote the $i$-th block that is separated according to the image resolutions. During training, the weights of the FP model remain constant, whereas the weights of the LP model, represented by $\{\theta_i\}_{i\leq n}$, are trainable.
We define the $k$-th mixed-precision (MP) model with implementing the block replacement starting from $k$-th blocks (between $\{Q_i \}_{i>k}$ and $\{F_i\}_{i>k}$), as: 
\begin{equation}
    M^k=\{Q_1,…,Q_k,F_{k+1},…,F_n\}
\end{equation}
where $M^k$ comprises the first $k$ blocks of LP models with trainable weights $\{\theta_i\}_{i\leq k}$ . It is notable to remark that the implementation of MP within the framework is implicit. As depicted in Fig. \ref{fig2}, in the computing graph, the LP forward flow is reused for MP flows in consideration of the calculation consistency of the $M^k$ and $Q$ in the first $k$ blocks.

\subsection{Training Objectives}
Given the input $x$ with task target $y$, the prediction of each branch is computed as:
\begin{equation}
\begin{aligned}
    &y_Q =Q(x;\{\theta_i\}_{i\leq n})\\
    &y_{M^k} =M^k(x;\{\theta_i\}_{i\leq k})\\
    &y_F =F(x)
\end{aligned}
\end{equation} 
The LP model's output denoted as $y_Q$, represents the final training result of our framework. $y_F$ and $y_{M^k}$ are employed as soft labels during QAT training to aid in directing the training of LP blocks.
As depicted in Fig. \ref{fig2}, the training targets are divided into two components: task targets $y$, and soft labels $y_F$, $y_{M^k}$. 
\begin{algorithm}[t]
    \SetAlgoLined
	\caption{BWRF Training w.r.t the low-precision backbone network $Q$ and the full-precision counterpart network $F$.}
	\label{alg:BWRF}
	\KwIn{Mini-batch $\{x,y\}$; trainable weights $\{\theta_i\}_{i\leq n}$ within the low-precision blocks $\{Q_i\}_{i\leq n}$; the full-precision blocks $\{F_i\}_{i\leq n}$.} 
    \KwOut {Updated weights $\{\theta_i\}_{i\leq n}$.} 
	\BlankLine

    Compute the FP output $y_{F}=F(x)$\;
    
    Initialize $x_{Q_0}=x$ \;
    
    Continuously compute the intermediate features via the LP blocks $\{x_{Q_i}|x_{Q_i}=Q_{i}(x_{Q_{i-1}};\theta_i)\}_{i\in [1,n]}$ \;

    Obtain the LP output $y_Q = x_{Q_n}$ \;

    \For{$k=1$ to $n-1$}{
        Initialize $x_{M^k_k}=x_{Q_k}$ \;
        
        Continuously compute the intermediate features via the FP blocks $\{x_{M^k_i}|x_{M^k_i}=F_{i}(x_{M^k_{i-1}})\}_{i\in [k+1,n]}$ \;

        Obtain the MP output $y_{M^k} = x_{M^k_n}$
    }

    Compute the loss $L$ according to Eq. \ref{eq:loss-target} - Eq. \ref{eq:loss} \;

    Propagate the gradients for LP weights $\dfrac{\partial L}{\partial \theta_i}=\dfrac{\partial L}{\partial x_{Q_{i}}} \dfrac{\partial x_{Q_{i}}}{\partial \theta_i}
    =\Big( \dfrac{\partial L}{\partial x_{Q_{i+1}}}\dfrac{\partial x_{Q_{i+1}}}{\partial x_{Q_{i}}} + \dfrac{\partial L}{\partial x_{M^{i}_{i+1}}}\dfrac{\partial x_{M^{i}_{i+1}}}{\partial x_{Q_{i}}} \Big)
    \dfrac{\partial x_{Q_{i}}}{\partial \theta_i}$    \;
    Update $\{\theta_i\}_{i\leq n}$ based on gradients.


\end{algorithm}
The training objective designated as $L_{target}$, which pertains to the classification target, is applied uniformly to both $Q$ and $M$:
\begin{equation}
    L_{target}=L_{ce}(y_Q,y)+\sum_{k=1}^{k<n} \alpha_k \cdot L_{ce}(y_{M^k},y)
    \label{eq:loss-target}
\end{equation}

\begin{table*}[tb]
    \centering
    \caption{\textbf{Comparison of validation accuracy on ImageNet using the ResNet-18 architecture.} The weights and activations are quantized to the same precision throughout 4-, 3-, and 2-bits. QAT methods are categorized based on uniform and non-uniform quantization manners. LSQ$^\dag$ denotes replicated benchmarks in accordance with our experimental setting for unbiased comparisons. The performance gain achieved with BWRF is denoted by $\Delta$.}
    \resizebox{0.9\textwidth}{!}{
    \begin{tabular}{ccc|cc|cc|cc}
     \toprule
     \multirow{2}{*}[-1.5pt]{Network} &\multirow{2}{*}[-1.5pt]{Manner} &\multirow{2}{*}[-1.5pt]{Method}  
  &\multicolumn{2}{c|}{4 bits} &\multicolumn{2}{c|}{3 bits} &\multicolumn{2}{c}{2 bits} \\
    \cmidrule{4-9}
  && &Top1 &Top5 &Top1 &Top5 &Top1 &Top5 \\
    \midrule
    \multirow{16}{*}{\thead{ResNet18 \\ FP(72.26)}} &\multirow{6}{*}{Non-uniform} &LQ-Nets &69.3 &88.8 &68.2 &87.9 &64.9 &85.9 \\
    &&QIL &70.1 &- &69.2 &- &65.7 &- \\
    &&DAQ &70.5 &- &69.6 &- &66.9 &- \\
    &&APOT &70.7 &89.6 &69.9 &89.2 &67.3 &87.5 \\
    &&LCQ &71.5 &- &70.6 &- &68.9 &- \\

\cmidrule{2-9}
    &\multirow{12}{*}{Uniform} 
    &PACT &69.2 &89 & 68.1 &88.2 &64.4 &85.6 \\
    &&DoReFa-Net &68.1 &88.1 &67.5 &87.6 &64.7 &84.4  \\
    &&DSQ &69.6 &88.9 &68.7 &- &65.2 &- \\
    &&LSQ &71.1 &90.0 &70.2 &89.4 &67.6 &87.6 \\
    &&LSQ+ &70.8 &- &69.3 &- &66.8 &- \\
    &&EWGS &70.6 &- &69.7 &- &67.0 &- \\
    &&BR &70.8 &89.6 &69.9 &89.1 &67.2 &87.3 \\
    &&QKD &71.4 &90.3 &70.2 &89.9 &67.4 &87.5 \\
    &&QFD &71.1 &89.8 &70.3 &89.4 &67.6 &87.8 \\

\cmidrule{3-9}
&&LSQ$^\dag$ &70.4 &89.4 &69.3 &88.8 &65.3 &86.2\\
    &&BWRF (ours) &\bf{71.9} &\textbf{90.5} &\textbf{70.8} &\textbf{89.9} &\textbf{67.7} &\textbf{87.9} \\
    &&$\Delta$  &+1.5 &+1.1  &+1.5 &+1.1 &+2.4 &+1.7 \\
    \bottomrule
    \end{tabular}}
    \label{tab:1}
\end{table*}

where $L_{ce}$ represents the cross-entropy for the classification task, $\alpha_k$ are the hyper-parameters utilized to balance the losses of FP branches. MP models are expected to perform the same functions as the LP backbone, wherein the alignment between the end LP blocks and their FP branches (e.g., $\{Q_{k+1},...,Q_n\}$ and $\{F_{k+1},...,F_n\}$) is implicitly applied, and the first LP blocks (e.g., $\{Q_1,...,Q_k\}$) are necessary to extract functional features applicable to both the FP and LP branches. The knowledge distillation objectives, represented as $L_{distill}$, are designed to enhance the performance of the LP and MP models. This is achieved through the utilization of soft label objectives, which are determined by the KL distance, $L_{kd}$, in accordance with the vanilla logit-based KD method \cite{hinton2015distilling}:
\begin{equation}
\begin{aligned}
   \label{eq:loss-distill}
    & L_{distill}=L_{kd}(y_Q,y_F )+L_{kd}(y_Q,y_{M_{n-1}}^{avg})\\
    & +\sum_{k=1}^{k<n} \alpha_k \Big(L_{kd} (y_{M^k},y_F)+L_{kd}(y_{M^k},y_{M_{k-1}}^{avg}) \Big)
\end{aligned}
\end{equation}
where the $y_{M^k}^{avg}$reflects the ensemble of preceding MP models and is computed using the average of the previous prediction:
\begin{equation}
   \label{eq:soft-label}
    y_{M^k}^{avg}=\frac{1}{k+1} \Big( y_{F} + \sum_{j=1}^{j\leq k} y_{M^j} \Big)
\end{equation}

Self-distillation \cite{zhang2019your} is the underlying principle of these designs, which facilitates the constructive mutual influence of the various self-components and is supported by empirical results.

The framework is then trained under the integrated objectives, as shown in Fig. \ref{fig2}:
\begin{equation}
    L= L_{target}+ L_{distill}
    \label{eq:loss}
\end{equation}

\section{Experiments}
In this section, we evaluate the effectiveness of the proposed framework compared to alternative QAT methods. We focus on quantization performance under the low-bit widths of 4/3/2. In this case, the issue of performance degradation resulting from quantizers becomes apparent and deserves particular consideration. 
The primary outcomes of the proposed method, along with a comparison to alternative QAT methods, will be plainly outlined in Section 4.2.
A comprehensive description of the ablation studies for the method will be provided in Section 4.3. In Section 4.4, the additional influence of the proposed method on QAT training will be analyzed by incorporating the visualization results.

\subsection{Experimental Setup}

\noindent
\textbf{Datasets and Networks.} We perform experiments on two standard image classification datasets: ImageNet \cite{russakovsky2015imagenet} and CIFAR-10 \cite{krizhevsky2009learning}. ImageNet contains about 1.2 million training and 50K validation images of 1,000 object categories. CIFAR10 contains 60,000 32x32 color images in 10 classes, with 6,000 images per class. 
In our experiments, images from the ImageNet training set were randomly cropped to 224×224 pixels and randomly flipped for data augmentation. For the validation set, images were first resized to 256×256 pixels, then a 224×224 center crop was taken to evaluate consistent image sizes. All images in the dataset underwent normalization. 
Similar preprocessing was done for CIFAR-10 as on ImageNet.
We evaluated our method in the Reset series \cite{he_deep_2016}, including ResNet18, ResNet20, ResNet34, ResNet50 and ResNet56. Among them, ResNet18, ResNet34 and ResNet50 were used for ImageNet, while ResNet20 and ResNet56 were used for CIFAR10. We inserted quantizers for weights or activations before convolution operators in each layer to perform quantization. 8-bit quantization was applied in the first and last layers.

\begin{table*}[tb]
    \centering
    \caption{\textbf{Comparison of validation accuracy on ImageNet using the ResNet-34 and ResNet-50  architectures.} The weights and activations are quantized to the same precision throughout 4-, 3-, and 2-bits. QAT methods are categorized based on uniform and non-uniform quantization manners. }
    \resizebox{0.9\textwidth}{!}{
    \begin{tabular}{ccc|cc|cc|cc}
\toprule
\multirow{2}{*}[-1.5pt]{Network}  &\multirow{2}{*}[-1.5pt]{Manner} &\multirow{2}{*}[-1.5pt]{Method} 
    &\multicolumn{2}{c|}{4 bits} &\multicolumn{2}{c|}{3 bits} &\multicolumn{2}{c}{2 bits} \\
    \cmidrule{4-9}
    && &Top1 &Top5 &Top1 &Top5 &Top1 &Top5 \\
\midrule
    \multirow{11}{*}{\thead{ResNet34 \\ FP(76.32)}} &\multirow{5}{*}{Non-uniform}  &LQ-Nets &- &- &71.9 &90.2 &69.8 &89.1 \\
    &&QIL &73.7 &- &73.1 &- &70.6 &- \\
    &&APOT &73.8 &91.6 &73.4 &91.1 &70.9 &89.7 \\  
    &&LCQ &74.3 &- &74.0 &- &72.7 &- \\
\cmidrule{2-9}
    &\multirow{6}{*}{Uniform} 
    &DSQ  &72.8 &- &72.5 &- &70.0 &- \\
    &&LSQ &74.1 &91.7 &73.4 &91.4 &71.6 &90.3 \\
    &&EWGS &73.9 &- &73.3 &- &71.4 &- \\
    &&QKD &74.6 &92.1 &73.9 &91.4 &71.6 &90.3 \\
    &&QFD &74.7 &92.3 &73.9 &91.7 &71.7 &90.4 \\
\cmidrule{3-9}    
    &&BWRF (ours) &\textbf{75.9} &\textbf{92.6} &\textbf{74.3} &\textbf{91.7} &\textbf{71.7} &\textbf{90.4} \\
    \midrule
    \multirow{9}{*}{\thead{ResNet50 \\ FP(80.11)}} &\multirow{4}{*}{Non-uniform}  &LQ-Nets &75.1 &92.4 &74.2 &91.6 &71.5 &90.3 \\
    &&APOT &76.6 &93.1 &75.8 &92.7 &73.4 &91.4 \\   
    &&LCQ &76.6 &- &76.3 &- &75.1 &- \\
\cmidrule{2-9} 
    &\multirow{5}{*}{Uniform} 
    &PACT &76.5 &93.2 &75.3 &92.6 &72.2 &90.5 \\
    &&DeReFa-Nets &71.4 &89.8 &69.9 &89.2 &67.1 &87.3 \\
    &&LSQ &76.7 &93.2 &75.8 &92.7 &73.7 &91.5 \\
    &&QKD &77.3 &93.6 &76.4 &93.2 &73.9 &91.6 \\
\cmidrule{3-9}
    &&BWRF (ours) &\textbf{79.0} &\textbf{94.2} &\textbf{77.8} &\textbf{93.5} &\textbf{74.2} & \textbf{91.6} \\
    \bottomrule
    \end{tabular}}
     
     \label{tab:2}
\end{table*}

\noindent
\textbf{Quantization Setting.} In accordance with the configurations established in prior QAT methods\cite{han2021improving,li2019additive,lee2021network,hubara2020improving}
, we implemented quantizers for activations and weights prior to convolution operators in each layer to simulate quantization. 
While the final fully-connected layer and the initial convolutional layer were both quantized to 8 bits, all other convolutional layers were quantized to a consistent low-bit width.
The weights of the low-precision network are consistently initialized using the pre-trained, full-precision counterparts. 
Training details are appended in the supplement.

\noindent
\textbf{Training Details.}
We use SGD with a momentum of 0.9 for all cases. On imageNet, the batch size was 1024 for all bit widths for ResNet18. For 4-bit quantization of ResNet18, the initial learning rate was set to 1e-2 with a weight decay of 1e-4. For 3-bit ResNet18, the settings were the same. For 2-bit ResNet18, the learning rate was set to 4e-2 and weight decay to 2e-5. For ResNet34, the batch size was 1024 for all except 2-bit. The 4-bit quantization used a learning rate of 2e-2 and weight decay of 1e-4. The 3-bit settings used a learning rate of 1e-2 and weight decay of 1e-4. The 2-bit ResNet34 used a learning rate of 2e-2, batch size of 512, and weight decay of 2e-5. For ResNet50, the batch size was 512 for all bit widths. The 4-bit quantization used a learning rate of 2e-2 and weight decay of 1e-4. The 3-bit settings used a learning rate of 1e-2 and weight decay of 1e-4. The 2-bit ResNet50 used a learning rate of 2e-2 and weight decay of 2e-5. Models were all trained for 120 epochs with the learning rate divided by 10 every 30 epochs.
For FP counterparts, we trained ResNet18 with knowledge distillation and applied pre-trained ResNet-34 and ResNet-50 from TIMM \cite{wightman2110resnet}. On CIFAR-10, the initial learning rate was set consistently to 4e-2 for 2, 3, and 4-bit levels with a weight decay of 1e-4 for 3/4-bit and 3e-5 for 2-bit. Models were trained for 300 epochs with the multistep schedule, dividing the learning rate by 10 at the 150th and 225th epochs.

\subsection{Main Results}
\noindent
\textbf{Results on ImageNet.} The performance evaluation of the proposed methods is presented on the ImageNet dataset, utilizing the ResNet-18, ResNet-34, and ResNet-50 architectures, respectively. The obtained results are summarized in Tab. \ref{tab:1} and Tab. \ref{tab:2}, which show the performance of our BWRF in the presence of low precision bit widths. "LSQ$^\dag$" denotes the outcomes achieved via the preliminary implementation outlined in section 3.1, while combined with the configurations specified in section 4.1. These results formed the baseline for the subsequent developments incorporated into our framework.
Our proposed framework consistently outperforms other QAT methods with respect to R18, R34, and R50.
Significantly, our framework employing uniform quantization surpasses the state-of-the-art results of non-uniform quantization techniques in almost all cases. This indicates that the framework offers a dependable approach to achieve uniform quantization, which is particularly well-suited for inferring fixed-points computation.

\noindent
\textbf{Results on CIFAR-10.} 
Tab. \ref{tab:3} presents the performance on CIFAR-10, showcasing a consistent enhancement in comparison to alternative QAT methods. The experimental findings suggest that BWRF can effectively convert the FP model to 3/4 bit with minimal performance degradation. 
It is worth mentioning that the BWRF model exhibits superior performance when operating at 4 bits compared to the FP model. This can be attributed to the utilization of mechanisms resembling self-supervision and auxiliary supervision, which will be elaborated upon in Section 4.3 following ablation experiments.

\begin{table}[tb]
    \centering
    \caption{\textbf{Comparison of validation accuracy on CIFAR-10 using ResNet-20 and ResNet-56.} LSQ$^\dag$ denotes replicated benchmarks in accordance with our experimental setting for unbiased comparisons.}
    \resizebox{0.5\textwidth}{!}{
    \begin{tabular}{cc|ccc}
    \toprule
         \multirow{2}{*}[-1.5pt]{Network} &\multirow{2}{*}[-1.5pt]{Method} &\multicolumn{3}{c}{Accuracy (\%)} \\
    \cmidrule{3-5}
         & &4 bits &3 bits &2 bits\\
    \midrule
    \multirow{8}{*}{\thead{ResNet20 \\ FP(92.96)}}  &DoReFa-Net &90.5 &89.9 &88.2 \\
    &PACT &91.7 &91.1 &89.7 \\
    &LQ-Net &- &91.6 &90.2 \\
    &PACT+SAWB+fpsc &- &- &90.5\\
    &QKD &93.1 &92.7 &90.5 \\
    &APoT &92.3 &92.2 &\textbf{91.0} \\  
    &LSQ$^\dag$ &92.67 &92.34 &90.74 \\
    &BWRF (ours) &\bf{93.13} &\bf{92.76} &90.84 \\
    \midrule
    \multirow{4}{*}{\thead{ResNet56 \\ FP(94.46)}}  &PACT+SAWB+fpsc &- &- &92.5 \\
    &APoT &94.0 &93.9 &92.9 \\
    &LSQ$^\dag$ &93.74 &93.69 &92.84 \\
    &BWRF (ours) &\bf{94.69} &\bf{94.31} &\bf{93.03} \\
    \bottomrule
    
    \end{tabular}
    }
    \label{tab:3}
\end{table}

\begin{table}[tb]
    \centering
    \caption{\textbf{Ablation study of training objectives.} The 4-bit results are obtained using  ResNet-20 and ResNet-56 on CIAFR-10.}
    \begin{tabular}{ccccccc}
    \toprule
         $y_Q$ &\multicolumn{1}{c|}{$y_{M_k}$} &\multicolumn{1}{r}{$y_Q$} &$y_{M_k}$
          &\multicolumn{1}{|c|}{\multirow{2}{*}[-0.8pt]{w/ $y^{avg}$}} &\multicolumn{2}{c}{Accuracy (\%)} \\
    \cmidrule{1-2} \cmidrule{3-4} \cmidrule{6-7}
   \multicolumn{2}{c|}{w/ target $y$}  &\multicolumn{2}{c|}{w/ FP label $y_F$} &\multicolumn{1}{c|}{}  &R20 &R56 \\
    \midrule
        \ding{51} &&&&&92.67&93.74 \\
        \ding{51} &\ding{51}&&&&92.89&94.02 \\
        \ding{51} &&\ding{51}&&&92.91&94.23 \\
        \ding{51} &\ding{51}&\ding{51}&&&93.02&94.35 \\
        \ding{51} &\ding{51}&\ding{51}&\ding{51}&&93.09&94.54 \\
        \ding{51} &\ding{51}&\ding{51}&\ding{51}&\ding{51}&93.13&94.69 \\
    \bottomrule
    \end{tabular}
    \label{tab:4}
\end{table}

\begin{table}[t]
    \centering
    \caption{\textbf{Ablation study of pruning MP models.} The 4-bit results are obtained using ResNet-20 and ResNet-56 on CIAFR-10.}
    \begin{tabular}{c|c|c|c|c}
    \toprule
        Network &base &w/ $M^1$ & w/ $M^2$  & w/ $M^1$ \& $M^2$ \\
        \midrule
        ResNet20  &92.67 &92.97 &93.02 &93.13 \\
        \midrule
        ResNet56  &93.74 &94.42 &94.56 &94.69 \\
    \bottomrule
    \end{tabular}
    \label{tab:5}
\end{table}

\begin{table}[tb]
    \centering
    \caption{\textbf{Extension to non-uniform quantization.} The results are obtained using ResNet-20 and ResNet-56 on CIAFR-10.}
    \begin{tabular}{cc|ccc}
    \toprule
         Network &Manner &4 bits &3 bits &2 bits \\
    \midrule
         \multirow{2}{*}[-1pt]{ResNet20} &uniform  &93.13 &92.76 &90.84 \\
    \cmidrule{2-5}
         &Non-uniform &93.16 &92.82 &91.06 \\
    \midrule
        \multirow{2}{*}[-1pt]{ResNet56} &uniform  &94.69 &94.31 &93.03 \\
    \cmidrule{2-5}
         &Non-uniform &94.74 &94.38 &93.14 \\
    \bottomrule        
    \end{tabular}
    \label{tab:6}
\end{table}

\subsection{Ablation Study}
\noindent
\textbf{Ablation of training objectives.} The results of the ablation experiment regarding the training objectives of BWRF are presented in Tab. \ref{tab:4}. The foundational implementation of QAT is training exclusively with $L_{ce} (y_Q,y)$; training with both $L_{kd}(y_Q,y_F)$ and $L_{ce} (y_Q,y)$ is analogous to distilling LP using the FP model. We gradually integrate training objectives posterior to section 3.2 as auxiliary supervision. 
The overall performance continues to improve with the incorporation of additional loss sources, indicating that these loss sources are effective for the auxiliary supervision of QAT. It is important to highlight that the structure implemented utilizing $L_{ce} (y_{M_k},y)$ is similar to the methodology employed to implement multi-exit for classification in DSN \cite{lee2015deeply}. Nevertheless, the branch that BWRF employs remains consistent, and the FP model verifies that the branch utilized as the classifier is rational. 
Furthermore, the implementation carried out utilizing MP outputs may be regarded as a unique form of self-supervised methodology \cite{zhang2019your} that integrates untrained branches for the purpose of extracting intermediate features; the effectiveness of this approach is assessed via empirical experiments. 
Overall, the integration of these auxiliary training objectives improves the quality of the QAT training framework and can be further integrated to yield better outcomes.

\noindent
\textbf{Ablation of MP branches.} The outcomes of ablation for MP branches are presented in Tab. \ref{tab:5}, which illustrates the beneficial effects of setting MP branches within the framework. As the branches are pruned, the performance returns to its initial state. 
It is noteworthy to mention that $M^2$ provides enhanced auxiliary supervision when compared to $M^1$, which implies that the replacement between intermediate blocks provides more reliable support.

\subsection{Analysis and Discussion}

\begin{figure}[t]
\centering
\includegraphics[width=1.0\columnwidth]{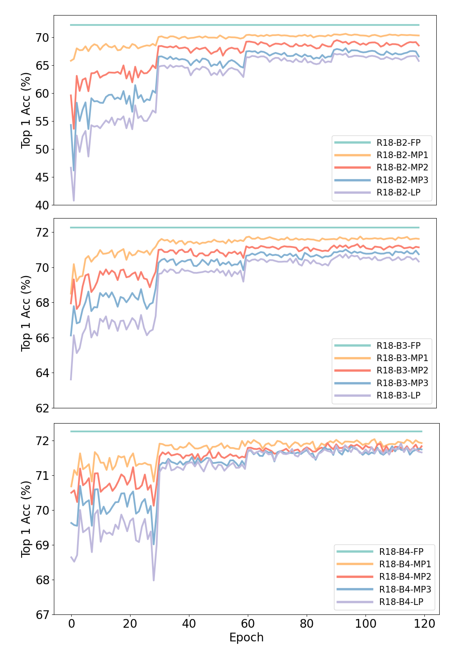} 
\caption{\textbf{Validation accuracy of mixed-precision models and low-precision backbone during training.} The results are obtained by ResNet-18 on ImageNet.}
\label{fig1e}
\end{figure}

\noindent
\textbf{Extension to non-uniform quantization.} The adaptability of BWRF to the non-uniform design was assessed, as illustrated in Tab. \ref{tab:6}. It is not unexpected that performance can be marginally enhanced by combining BWRF with a non-uniform quantizer, considering that non-uniform quantization offers a more reliable representation at lower levels of bit widths. Nevertheless, upon comparison, the distinction between uniform and non-uniform is not particularly significant. For increased adaptability, we suggest implementing uniform quant in accordance with BWRF training.

\begin{figure}[t]
\centering
\includegraphics[width=1.0\columnwidth]{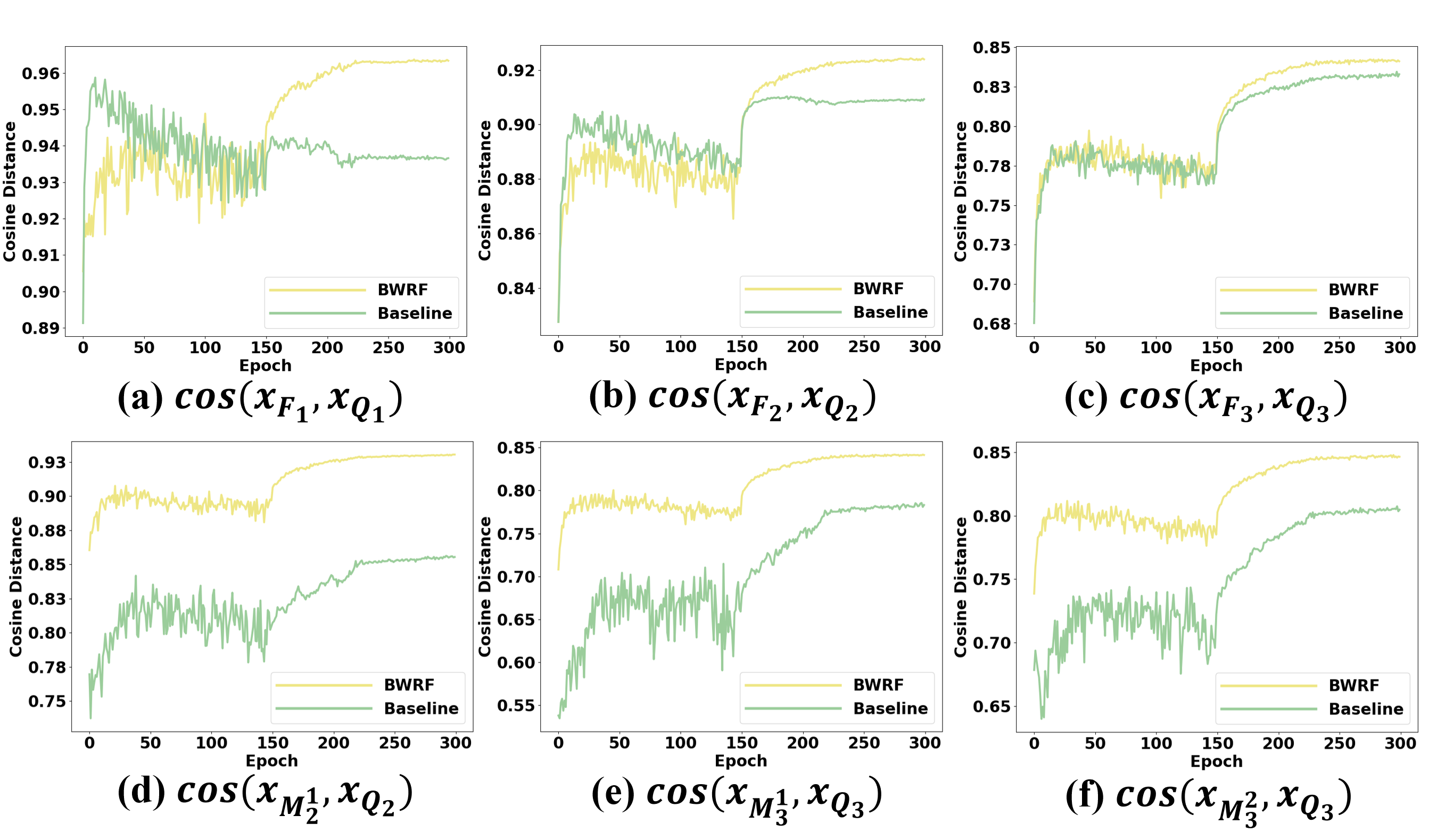} 
\caption{\textbf{Measures of feature similarity.} The results of cosine distances are obtained by ResNet-18 on CIFAR-10 under 4-bit quantization.}
\label{fig2e}
\end{figure}

\noindent
\textbf{Analysis of MP performance.} An illustration of the performance trajectory of intermediate mixed-precision models throughout the training procedure can be found in Fig. \ref{fig1e}. MP generally exhibits performance that lies between that of FP and LP, with a gradual deterioration noted as the number of LP blocks increases. This observation aligns with the expectations that were held. LP inevitably encounters a decrease in performance due to the representation capacity limitation that occurs during the phased block replacements from the FP block to the LP block. This phenomenon is at its most conspicuous at two and three bits. It is worth mentioning that the differentiation between the branches of the 4-bit model does not become immediately evident during the late training phase. The discrepancy between the FP and the first MP appears to be the most significant obstacle. Therefore, to enhance the accuracy of ancillary supervision, an intuitive optimization technique involves further dividing the first LP block into several subblocks.
\begin{figure}[t]
\centering
\includegraphics[width=1.0\columnwidth]{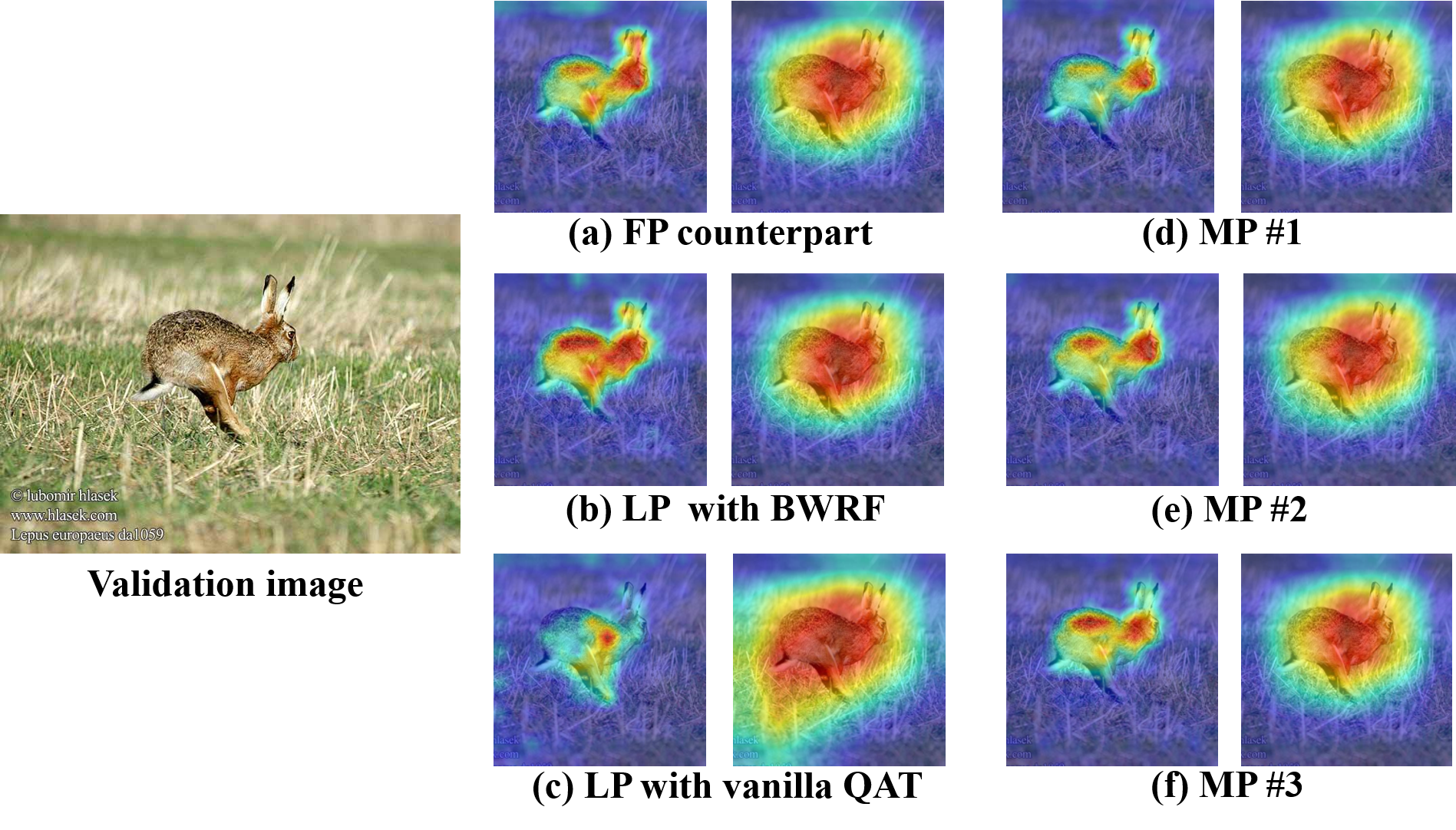} 
\caption{\textbf{Visualization results of class activation mapping (CAM).} Two visualized targets are established on the third and fourth blocks' output layers. (a-b) The results of the full-precision model $F$ and low-precision model $Q$ trained with BWRF. (c) The results of the vanilla model with baseline implementation. (d-f) The results of mixed-precision models $M^1$, $M^2$, $M^3$, respectively.}
\label{fig3e}
\end{figure}

\noindent
\textbf{Analysis on distance between intermediate features.} As shown in Fig. \ref{fig2e}, to obtain insights regarding the comparison between quantized and full-precision features in the QAT process, the cosine distance between intermediate features of LP and MP/FP is computed throughout the training phase. A comparative analysis of the situation is performed utilizing the initial LSQ method. The degree of alignment between the features of FP and LP is illustrated in Fig. \ref{fig2e} (a)-(c), which depicts the direct correspondence between the two designs.
The cosine distance of the LP model undergoing BWRF training approaches 1 during the late training phases, indicating a more optimal fit with FP. This finding provides empirical validation for the hypothesis that the auxiliary supervision implemented in the BWRF configuration can assist the LP model in incorporating the previous FP representation into the LP features.
The distance relationship between LP and MP features is illustrated in Fig. \ref{fig2e} (d)-(f). This relationship can be used to determine the degree of block-wise similarity between FP and LP, thereby offering a more comprehensive view of the fitting situation between FP and LP. As an example, Fig. \ref{fig2e} (d) illustrates the value of $cos(x_{Q_2},x_{M^1_2})$, which quantifies the separation between ${Q_2}(x_{Q_1})$ and ${F_2}(x_{Q_1})$ in order to represent the matching distance between $Q_2$ and $F_2$ with respect to a given $Q_1$ block.  In block-wise distance detection, BWRF is invariably greater than the baseline method; therefore, as expected, fitting LP and FP via block replacement is valid. 
Additionally, it is critical to note that the model prioritizes the fitting of block-wise representations Fig. \ref{fig2e} (d)-(f) when the block replacement setting is enabled. Consequently, the initial depiction of the overall features of FP may be marginally delayed Fig. \ref{fig2e} (a)-(b); however, it possesses the capability to converge towards higher values gradually.

\noindent
\textbf{Visualization with class activation mapping.} Fig. \ref{fig3e} shows the Class Activation Mapping (CAM) generated by the Layer-CAM method \cite{jiang2021layercam}. 
The discovery that the LP, when applied to BWRF, yields outcomes similar to those of the FP counterpart, indicates that the LP's feature extraction and gradient backpropagation computation align with those of the FP, thus validating the LP's ability to effectively convey the FP representation. 
The progressive step results from FP to LP are displayed across MPs' results, illustrating that continuous and logical intermediate step models can be derived from FP to LP through the sequential block-wise replacement process.

\section{Conclusion}
In this paper, we propose the block-wise replacement framework (BWRF) for quantization-aware training (QAT) by applying auxiliary supervision for low-bit models via the FP counterpart in order to mitigate the representation limitation and gradient mismatch issues that arise from the incorporation of discrete quantizers.
We analyze and discuss the latent insights of the proposed framework through empirical experiments.
To be emphasized, we obtain state-of-the-art results with uniform quantization settings for low-bit widths of 4/3/2, and we even outperform non-uniform methods in most instances.
The framework itself is neat and flexible, necessitating only a concise wrapper for code implementations.
We believe that BWRF can facilitate network quantization by providing extensions to QAT methods.

\bibliographystyle{plainnat} 
\bibliography{IEEEabrv,reference}

\end{document}